\pdfoutput=1
\documentclass[11pt,a4paper]{article}
\usepackage[hyperref]{emnlp2020}
\usepackage{times}
\usepackage{latexsym}

\usepackage{graphicx}
\usepackage{multirow}
\usepackage{csquotes}

\usepackage{microtype}

\aclfinalcopy 

\title{``Did you really mean what you said?" : Sarcasm Detection in Hindi-English Code-Mixed Data using Bilingual Word Embeddings}

\author{Akshita Aggarwal, Anshul Wadhawan, Anshima Chaudhary and Kavita Maurya \\
  Department of Computer Engineering\\
  Netaji Subhas University of Technology\\
  Dwarka, New Delhi \\
  \texttt{\{akshitaa, anshulw, anshimac, kavitam\}.co.16@nsit.net.in} \\}
\date{}

\begin{document}
\maketitle
\begin{abstract}
With the increased use of social media platforms by people across the world, many new interesting NLP problems have come into existence. One such being the detection of sarcasm in the social media texts. We present a corpus of tweets for training custom word embeddings and a Hinglish dataset labelled for sarcasm detection. We propose a deep learning based approach to address the issue of sarcasm detection in Hindi-English code mixed tweets using bilingual word embeddings derived from FastText and Word2Vec approaches. We experimented with various deep learning models, including CNNs, LSTMs, Bi-directional LSTMs (with and without attention). We were able to outperform all state-of-the-art performances with our deep learning models, with attention based Bi-directional LSTMs giving the best performance exhibiting an accuracy of 78.49\%.
\end{abstract}

\section{Introduction}

With the advent of social media, a large part of human interaction is carried out online. This leads to generation of huge amounts of textual data that can be used to draw meaningful inferences. Social media websites like Facebook, Twitter, Reddit etc are used by people across cultures to communicate with each other and voice their opinions.

With the large amount of data available from social media, the study of various types of linguistic expressions like irony, humor, sarcasm, aggression, hate etc has become a keen research area. Especially in the field of NLP, automatic detection of these expressions is being widely explored \cite{1}. Automatic detection involves using computational methods to detect the presence of a particular emotion.

Although English is the language most commonly used on these websites, a majority of people are not native English speakers. These people therefore prefer to communicate in languages other than English \cite{2}. A study on the languages that are most commonly used for exchange of information on Twitter showed that around 50\% of the posts are written in languages other than English \cite{3}. This raises the opportunity of dealing with multi-lingual data generated by the social media sites. Various statistics show that around 26\% of the Indian population is
bilingual\footnote{\url{https://en.wikipedia.org/wiki/Multilingualism\_in\_India}}. This gives rise to the phenomenon of code-switching and code-mixing  \cite{4}. Code mixing takes place when speakers use two or more languages below clause level in a single social context. Multilinguals use such a mixture of languages, particularly on social media \cite{5}. There are multiple challenges of working with code-mixed data like large amount of new constructions that are a result of combining lexicons and syntax of two different languages, availability of very small amounts of annotated data and use of very different approaches when compared to mono-lingual data \cite{6}.

\begin{figure*}[h!]
    \centering
    \includegraphics[width=13cm]{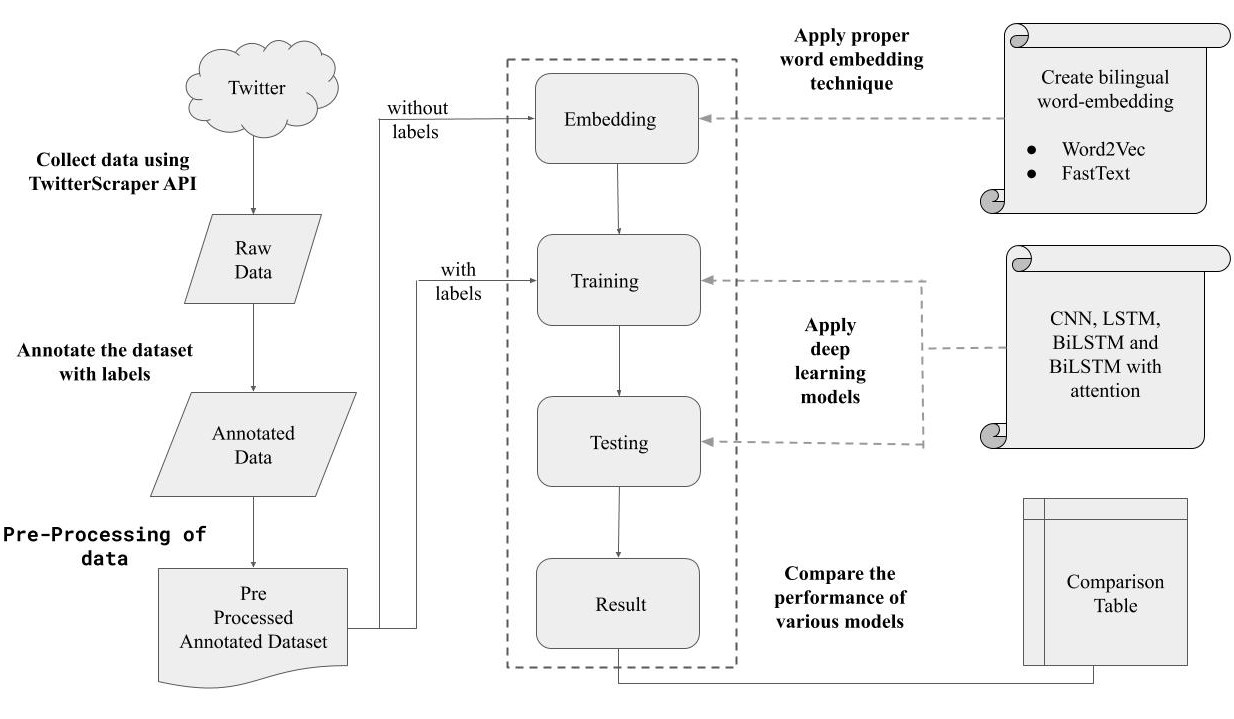}
    \caption{Proposed Methodology}
    \label{figure}
\end{figure*}

In this paper, we wish to work on detecting one of the most popular linguistic constructs used across social medias, ‘sarcasm’. The cambridge-dictionary\footnote{\url{https://dictionary.cambridge.org/}} defines sarcasm as ‘the use of remarks that clearly mean the opposite of what they say’. Example: "You have been working hard," he said with heavy sarcasm, as he looked at the empty page.

Starting with the earliest known work which focuses on sarcasm detection in speech \cite{7}, this domain has been widely explored in sentiment analysis. Since sarcasm is a sentiment, detection of sarcasm is important in order to predict the sentiment of a sentence. Being a challenging problem, automatic detection of sarcasm has been a popular area of research.

Although a lot of work has been carried out on sarcasm detection in English  \cite{8,9}, the detection of sarcasm in code-mixed language like Hinglish (Hindi-English) is relatively unexplored. The current state-of-art performance is proposed via a random forest model on a dataset of 5000 Hinglish tweets \cite{10}. 

The contributions of our work includes :
\begin{enumerate}
\item In this paper, we have experimented with deep-learning approaches to detect sarcasm in the Hindi-English code mixed dataset. The corpus prepared is released along with the paper.
\item  Deep learning is being used extensively in the domain of natural language processing and has given satisfactory results \cite{11}. In our work, we propose five different deep learning models namely, Series CNN, Parallel CNN, LSTM, Bi-directional LSTM and Bi-directional LSTM with attention.
\item The proposed models take self-trained bilingual word embeddings generated by Hindi-English code mixed data as input.
\item Our work present an alternate approach to the work done using traditional machine learning models like SVMs and random forests \cite{10}. 
\end{enumerate} 

\section{Proposed Methodology}


\subsection{Dataset Creation}

The current dataset provided in paper \cite{10} contains 5250 tweets, out of which 504 tweets are labelled as sarcastic while the remaining 4746 tweets are labelled as not sarcastic. All the deep learning models seemed to erroneously predict all tweets to be not sarcastic, since this dataset is highly skewed as well as insufficient. Therefore, to meet the model needs, we created a larger class-balanced dataset by scraping relevant tweets from twitter using TwitterScraper API\footnote{\url{https://github.com/taspinar/twitterscraper}} with search tags like \#sarcasm, \#irony, \#humor, \#bollywood, \#cricket along with some common hindi words to obtain Hinglish data. 

\subsection{Dataset Annotation and Analysis}

We were able to obtain around 427k tweets for training the proposed deep learning models. After carefully filtering out the obtained tweets for Hindi-English code mixed entries, we were successful in creating a corpus of 100k Hindi-English code mixed tweets with 49\% entries being sarcastic and remaining 51\% being non-sarcastic. The annotation scheme was based on the search tags(hashtags) used for scraping the tweets. We marked all examples fetched with hashtags like sarcasm, irony etc to have a positive sarcasm label, whereas all examples with generic hashtags like cricket, bollywood etc to have a negative sarcasm label. This annotation scheme was susceptible to noise, however, as a quality check measure, we manually traversed the data and noticed that the noisy examples were meagre in proportion. Also, the noisy examples were necessary for the models to generalize well on the diverse dataset we obtained. Having a class-balanced dataset was significant to our problem to ensure that deep-learning models learn the right trends, not being biased towards a particular class. Embeddings were initally trained on solely Hinglish data, which was later on added with English data. The embedding training dataset, labelled sarcasm detection dataset and the proposed deep learning classification models are made available online \footnote{\url{https://github.com/Akshitaag/Sarcasm_Detection}} to facilitate further research.

\begin{table}[]
\centering
\begin{tabular}{|l|l|}
\hline
{\color[HTML]{000000} \textbf{Category}} & {\color[HTML]{000000} \textbf{Tweet Count}}      \\ \hline
{\color[HTML]{000000} Total Tweets}      & {\color[HTML]{3C4043} 106899} \\ \hline
{\color[HTML]{000000} Sarcastic}         & {\color[HTML]{3C4043} 52587}  \\ \hline
{\color[HTML]{000000} Non-Sarcastic}     & {54312}                          \\ \hline
\end{tabular}
\caption{Tweets per category}
\label{table:1}
\end{table}

Examples of some annotated data :
\newline

Tweet: Koi Rah Mushkil Nahi hain bus vo rah \#bengalurutraffic se bach jaayein  \#sarcasm @random
\newline

Translation No path is difficult as long as it does not pass through Bangalore traffic.
(Bangalore is an Indian city infamous for it's traffic)
\newline
Sarcasm : YES
\newline

Tweet : Hindustan ke tamam log chahte h ke jis trah se auraton ke upar crime bhadr rha h gang rape ke waqia ho rha iske liye central govt wali modi sarkar 1 strong law bnaye
\newline

Translation: All indians want Modi government to make strong laws on crime against women
\newline
Sarcasm: NO

\subsection{Data Preprocessing}

The data obtained from social media is very noisy and a lot of preprocessing is required. While creating the dataset,  we removed the ‘\#’ symbols from the data, along with removing all the mentions (@). We also removed rare words (words having occurrence of less than 10 in the entire dataset) and search tags (like cricket, sarcasm) to avoid our deep learning models being biased towards certain words while learning. Further, URLs and punctuation marks were also removed.

\subsection{Creation of Hindi-English Bi-lingual Word Embeddings}

Being a text classification problem, it is essential for the words of the dataset to be first converted to vector representations. Word embedding is learned from unannotated plain text, useful in determining the context in which a given word is used. They provide a dense vector representation of syntactic or semantic aspects of a word \cite{12}.
To create a Hindi-English word embedding, we needed a huge amount of data. We used TwitterScraper API to extract 427k Hinglish tweets and 300k English tweets from Twitter for Hindi-English code-mixed data. For the Hinglish code mixed tweets, we removed the tweets obtained in pure Devnagri and kept only those which were a mixture of both Hindi and English sentences. The above obtained dataset was further processed to remove rare words, hashtags and mentions to obtain a less noisy corpus for training word embeddings.

We experimented with 2 different kinds of word embeddings for two types of datasets, one which solely consisted of Hinglish tweets, the other which consisted of 300k English along with Hinglish tweets. We chose to experiment with a mixture of Hinglish and English tweets in order to get the co-relations between the words of the two languages. Each of these variations, after similar processing (removing hashtags, URLs, punctuations, user mentions and keywords used for scraping), were tried for two types of embeddings:

\begin{figure*}[h!]
\begin{center}
    \centering
    \includegraphics[width=13cm]{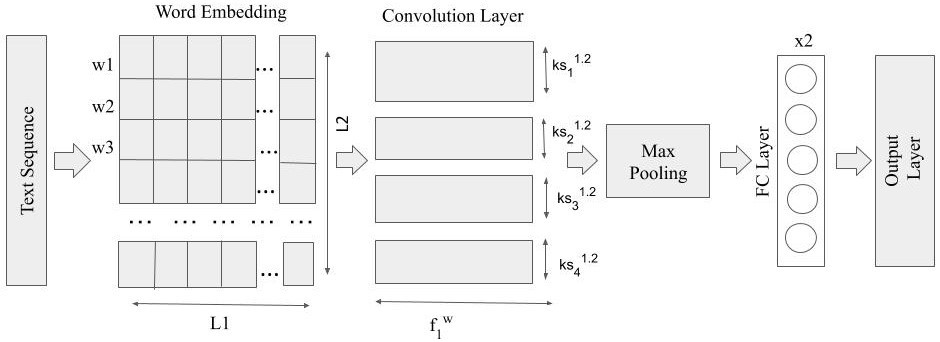}
    \caption{CNN Model 1.2 architecture}
    \label{figure}
\end{center}
\end{figure*}

{\bf Word2Vec}: In this embedding, words in the corpus are converted into vectors, where words that share common context are placed closed to each other in the vector-space \cite{13}. 
\newline Since Word2Vec is pre-trained for English dataset only, we had to train our model on custom Hindi-English code mixed dataset, to obtain Hinglish word embeddings.

{\bf FastText}: FastText which was given by Facebook in 2016,  is an addition to the Word2Vec embeddings  \cite{14}. Rather than giving individual words to a model, FastText breaks down the words into multiple sub-words, also known as n-grams \cite{15}. During the training of the model, weights are learned for all the n-grams along with the complete word. Unlike Word2vec, rare words can be appropriately featured as now it is much more likely that some of their n-grams also occur in other words. This is especially true for social media text where people use multiple spellings for the same words (amaze, amazeee, amazing, amazinggg).

\subsection{Deep Learning Models}

We propose 5 different models to experiment with the above problem. The models tested include Series CNN, Parallel CNN, LSTM, Bi-directional LSTM and Bi-directional LSTM with attention. Word embeddings, as generated by FastText and Word2Vec custom data trained representations, served as input to the models, generate as output a binary variable depicting the probability of the corresponding tweet being sarcastic.

\subsubsection{Convolutional Neural Networks (CNN)}

CNNs have the ability to extract features from data provided to it as input. In our case, the input data is a set of word vectors, over which convolution operation is performed to extract features and thereby, perform classification.

We propose 2 CNN deep learning model architectures, one which has convolution layers in series, denoted by model 1.1 and the other which has convolutional layers in parallel, denoted by model 1.2. 
Both the models have an embedding layer as the first one, which is used to select the word vector representations corresponding to the words of the tweet under consideration during the training session, from the word embedding matrix. In model 1.1, the embedding layer is followed by a couple sets of convolution and max pooling layers in series whereas in model 1.2, it is followed by 4 single dimensional convolution layers in parallel.


The convolution layer is responsible for the extraction of features from the word vectors provided as input. The outputs of these layers, concatenated in case of model 1.2, are fed to a global max pooling layer with a dropout activated. This layer is further followed by 3 dense fully connected layers, the final layer with a single neuron, which is responsible for the classification. We have used dropout in the the global max pooling layer so as to reduce overfitting, which already is low due to the large dataset. However, on its application, the difference between the validation accuracy and training accuracy reduced, also leading to better convergence.

\subsubsection{Recurrent Neural Networks (RNN)}

The meaning of a word depends on the context in which it is used. For example, \newline \newline
	Sentence 1 : We dined at a small Mexican restaurant and spent the meal discussing general topics.\newline
	Sentence 2 : General Zod is an enemy of Superman.\newline
	
The word general, in the above sentences, carries different meaning depending on the context in which it is used. Thus, in order to record the context of a particular word, i.e. the words surrounding the word under consideration, RNNs are used. There are different ways to capture the context of a particular word, each having its unique mechanism to model the meaning of the word depending on words coming before and after the word.

The RNN model equations and corresponding notation have been taken from \cite{18}. Given an input sequence x = (x\textsubscript{1}, x\textsubscript{2}, ..., x\textsubscript{t-1}, x\textsubscript{T}), the output vector sequence y = (y\textsubscript{1}, y\textsubscript{2}, ..., y\textsubscript{T-1}, y\textsubscript{T}) and hidden vector sequence h = (h\textsubscript{1}, h\textsubscript{2}, ..., h\textsubscript{T-1}, h\textsubscript{T}) are computed in a standard recurrent neural network by evaluating the below equations from t = 1 to t = T:\\ \\
\begin{math}
\begin{array}{l}
h_{t}=\mathcal{H}\left(W_{x h} x_{t}+W_{h h} h_{t-1}+b_{h}\right) \\ \\
y_{t}=W_{h y} h_{t}+b_{0} \\ \\
\end{array}
\end{math}
\\ where weight matrices are denoted by \begin{math} W \end{math} terms, bias vectors are denoted by \begin{math} b \end{math} terms, and hidden layer function is given by \begin{math} \mathcal{H} \end{math}.

{\bf Long Short-Term Memory (LSTM)}: LSTMs have been successfully applied to binary text classification problems like political text classification \cite{17}, by capturing the appropriate context. Also, the vanishing gradient problem in RNNs has been addressed successfully by LSTMs \cite{19}. The context of a word depends on the words occurring before the word under consideration. In order to model this scenario, an LSTM based network is constructed.The LSTM design comprises of a set of repetitively associated subnets, known as memory blocks. Each block contains at least one self-associated memory cells along with three multiplicative units - the input, output and forget gates - that give regular functionality of write, read and reset operations to the cells.\\ \\ \indent
A LSTM network is framed precisely like a basic RNN, other than the nonlinear units in the hidden layers being supplanted by memory blocks. The multiplicative gates permit LSTM memory cells to store and access data over extensive stretches of time, in this manner maintaining a strategic distance from the vanishing gradient issue. For instance, as long as the input gate stays shut ( has an activation near 0), the activation of the cell won't be overwritten by the new inputs showing up in the network, and can in this manner be made accessible to the net a lot later in the succession, by opening the output gate. This allows the LSTM network to carry forward semantic qualities of initial parts of the sentence to the later parts. In our architecture, an LSTM layer is appended to the embedding layer in turn followed by 2 dense fully connected layers. The final output layer has a single neuron carrying out the classification depending on the extracted context based features. LSTM blocks have the structure as shown in Figure 3, and are based on the equations presented below : 

\begin{figure}[h!]
\begin{center}
    \centering
    \includegraphics[width=6cm]{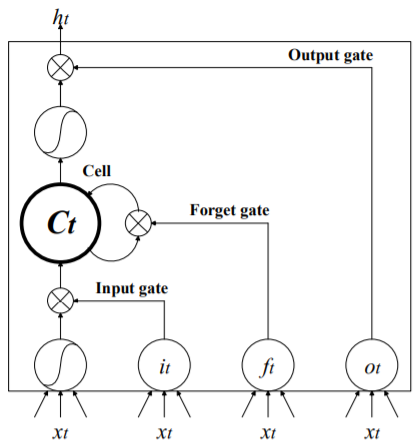}
    \caption{LSTM block structure}
    \label{figure}
\end{center}
\end{figure}

\begin{figure*}[h!]
\begin{center}
    \centering
    \includegraphics[width=12cm]{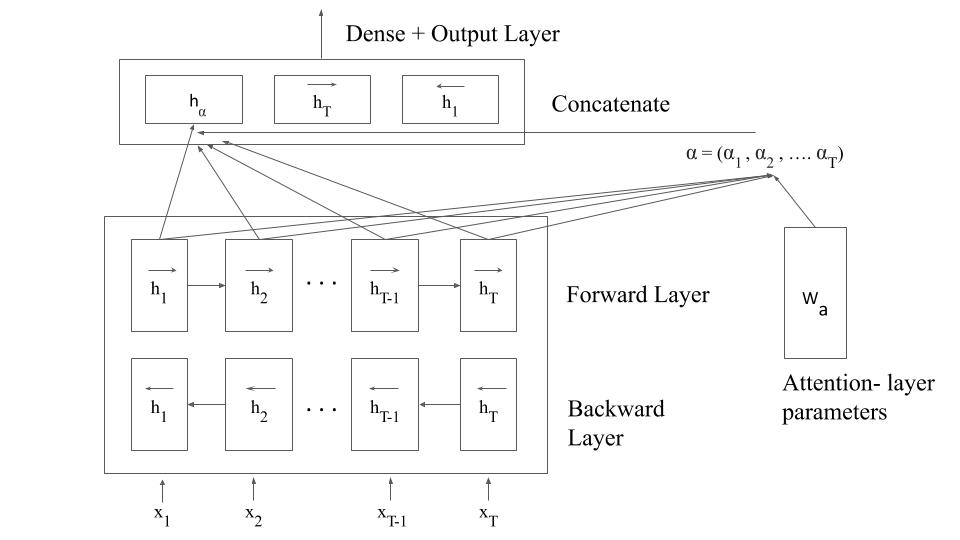}
    \caption{Attention based Bi-directional LSTM model architecture}
    \label{figure}
\end{center}
\end{figure*}

\noindent \begin{math}
\begin{array}{l}
i_{t} = \sigma\left(W_{x i} x_{t}+W_{h i} h_{t-1}+W_{c i} c_{t-1}+b_{i}\right) \\ \\
f_{t} = \sigma\left(W_{x f} x_{t}+W_{h f} h_{t-1}+W_{c f} c_{t-1}+b_{f}\right) \\ \\
\end{array}
\end{math}

\noindent \begin{math}
\begin{array}{l}
c_{t} = f_{t} c_{t-1}+i_{t} \tanh \left(W_{x c} x_{t}+W_{h c} h_{t-1}+b_{c}\right) \\ \\
\end{array}
\end{math}

\noindent \begin{math}
\begin{array}{l}
o_{t} = \sigma\left(W_{x o} x_{t}+W_{h o} h_{t-1}+W_{c o} c_{t}+b_{o}\right) \\ \\
h_{t} = o_{t} \tanh \left(c_{t}\right)
\end{array}
\end{math}
\newline

where logistic sigmoid function is denoted by \begin{math} \sigma \end{math}, the input gate, forget gate, output gate, and cell activation vectors are denoted by \begin{math} i, f, o \end{math} and \begin{math} c \end{math}, all having the same size as the hidden vector \begin{math} h \end{math}. The hidden-input gate matrix is represented by \begin{math} W_{h i} \end{math}, and the input-output gate matrix is represented by \begin{math} W_{x o} \end{math}.

{\bf Bi- directional LSTM}: Bi-directional LSTMs have been applied and proved to be successful in capturing the context for text classification tasks \cite{16}. The context of a word not only depends on the words occurring before it, but also on the words occurring after it. Modelling this requires memory cells in the backward direction which maintain the history of words along with cells in the forward direction for the words not yet explored. To achieve this and capture the composition semantics of Hindi-English code mixed data, two LSTM layers are appended to the input embedding layer. The output features from the two layers, after concatenation (\overrightarrow{h\textsubscript{t}},  \overleftarrow{h\textsubscript{1}}), are flattened and fed to 2 dense fully connected layers. The classification is performed by a single neuron, as in all other models. The BiLSTM computes the forward hidden sequence \begin{math} \overrightarrow{h_{t}} \end{math} by traversing the forward layer from t = 1 to T, the backward hidden sequence \begin{math} \overleftarrow{h_{t}} \end{math} by traversing the backward layer from t = T to 1, and updates the output \begin{math} y_{t} \end{math} as : 
\\ \\ \\
\begin{math}
\begin{array}{l}
\overrightarrow{h_{t}} =\mathcal{H}\left(W_{x\overrightarrow{h}} x_{t}+W_{\overrightarrow{h} \overrightarrow{h}} \vec{h}_{t+1}+b_{\overrightarrow{h}}\right) \\ \\
\overleftarrow{h_{t}} =\mathcal{H}\left(W_{x\overleftarrow{h}} x_{t} + W_{\overleftarrow{h} \overleftarrow{h}} \overleftarrow{h}_{t+1}+b_{\overleftarrow{h}}\right) \\ \\
y_{t} =W_{\overrightarrow{h} {y}} \overrightarrow{h}_{t}+W_{\overleftarrow{h}y} \overleftarrow{h}_{t}+b_{y}
\end{array}
\end{math}

{\bf Attention based Bi-directional LSTM}: Here, we propose a technique based on attention along with the bi-directional LSTM network. The attention based network focuses on filtering the noisy elements of a sentence by learning the words which cause the greatest effect towards deciding the final output (sarcastic or not sarcastic) of the sentence under consideration. While the bi-directional LSTM network uses concatenated (\overrightarrow{h\textsubscript{T}},  \overleftarrow{h\textsubscript{1}}) which is then fed to the dense layers, attention based bi-directional LSTM network is different in the concatenation process.  Along with the above state representations, (\overrightarrow{h\textsubscript{T}} denoting the final state representation in the forward direction and \overleftarrow{h\textsubscript{1}} denoting the first state representation in the backward direction), the attention based network inculcates the weighted summation, calculated by detecting the influence of each word, of all the time steps (denoted by  \overrightarrow{h\textsubscript{t}}, \overleftarrow{h\textsubscript{t}}). Thus, all these hidden states are concatenated and passed on to 2 dense fully connected layers. Final classification is performed by a single neuron, as usual.

\begin{table*}[h!]
\centering
\begin{tabular}{|l|l|}
\hline
{\color[HTML]{000000} \textbf{Traditional Models}}            & {\color[HTML]{000000} \textbf{Accuracy}} \\ \hline
{\color[HTML]{000000} Naive Bayes}                          & {\color[HTML]{212121} 54.17}             \\ \hline
{\color[HTML]{000000} Random Forest}                             & {\color[HTML]{212121} 63.37}            \\ \hline
{\color[HTML]{000000} Linear SVM} & {\color[HTML]{000000} 69.04 }                  \\ \hline
{\color[HTML]{000000} RBF Kernel SVM}                            & {\color[HTML]{212121} 71.23}             \\ \hline
\end{tabular}
\caption{Accuracy of ML models}
\label{table:2}
\end{table*}

\section{Experimental Settings}
\begin{table*}[]
\centering
\begin{tabular}{|l|l|l|l|l|}
\hline
\multicolumn{1}{|c|}{\multirow{2}{*}{\begin{tabular}[c]{@{}c@{}}\textbf{DL} \\ \textbf{Models}\end{tabular}}} & \multicolumn{2}{|c|}{\begin{tabular}[c]{@{}c@{}}\textbf{Hinglish}\\ \textbf{Data}\end{tabular}} & \multicolumn{2}{c|}{\begin{tabular}[c]{@{}c@{}}\textbf{Hinglish + English}\\ \textbf{Data}\end{tabular}} \\ \cline{2-5} 
\multicolumn{1}{|c|}{} & Word2Vec (a) & FastText (b) & Word2Vec (a) & FastText (b) \\ \hline
1.1 Series CNN & 72.86 & 72.65 & 74.09 & 73.51 \\ \hline
1.2 Parallel CNN & 74.28 & 73.41 & 75.00 & 74.32 \\ \hline
2.1 LSTM & 76.19 & 75.25 & 77.24 & 75.55 \\ \hline
2.2 Bi-LSTM & 77.12 & 76.25 & 78.28 & 77.12 \\ \hline
2.3 Attention Bi-LSTM & 78.19 & 77.11 & 78.40 & 78.06 \\ \hline
\end{tabular}
\caption{Accuracy of DL Models}
\label{tab:my-table}
\end{table*}

For the training sessions, we made a ten percent validation split and shuffled the training dataset, so that the model does not capture sequence trends, if any, in the training data, for a total of 20 epochs. The model checkpoints were saved at every epoch and those checkpoints which were saved before the model begins to overfit and the difference between the training and validation accuracies becomes significant, were used to calculate the accuracy numbers on the ten percent test dataset split. There are many important hyper-parameters in the training script of embeddings as well as the proposed models, which are tuned to produce the best training results on the validation data split. For training the word embeddings (both Word2Vec and FastText), we used an embedding size of 300, window length of 10 and negative sampling polarity. In all the models, adam optimizer along with binary cross entropy loss function has been used. All the layers have relu activation function with the exception of output layer having sigmoid activation function. We evaluated the performance of CNN models with different values for kernel\_size, number\_of\_kernels, dropouts and strides. The best results are obtained with the following values : 
\\ \\
stride = 1, number\_of\_kernels = 200, dropout = 0.5
ks\textsubscript{1}\textsuperscript{1.2} = 3, ks\textsubscript{2}\textsuperscript{1.2} = 6, ks\textsubscript{3}\textsuperscript{1.2} = 9,
\newline
ks\textsubscript{4}\textsuperscript{1.2} = 12,
ks\textsubscript{1}\textsuperscript{1.1} = 7
\\ \\
For all the proposed RNNs, the following hyper parameter combination is used : \\ \\ dropout\_for\_recurrent\_state = 0.2, 
\newline dropout\_for\_input\_state = 0.2, 
\newline number\_of\_LSTM\_units = 150
\\ \\We used the same hyper parameter values for models 2.2 and 2.3, as in model 2.1, so as to study the impact of imposing bidirectional nature to the LSTM layer, as well as exploring the effect of attention introduction. The parameters resulted in best outputs as confirmed later by trying out different values for the same.

\section{Results}

The dataset, as presented in \cite{10}, being insufficient and skewed for our deep learning model architectures, we ran the state-of-the-art models on our proposed dataset to carry out unbiased accuracy comparison of state-of-the-art techniques and neural network based models. 

The results for the same have been presented in Table 2. Using all features, the traditional state-of-the-art models: RBF kernel SVM, random forest and linear SVM, proposed the best accuracy of 71.23\% on the proposed corpus. We tested all the deep learning models with both Word2Vec and FastText based word representations. The results of both have been presented in Table 3 where model (a) and (b) refer to application of Word2Vec and FastText generated word embeddings respectively.

To the best of our knowledge, we are the first to implement and analyze deep learning model architectures and different word representations for detection of sarcasm in Hindi-English code-mixed data with a dataset large enough for deep learning models. All the proposed deep learning models performed better than the traditional state-of-the-art models, where the attention based Bi-directional LSTM network produced the best accuracy of 78.49\%. In Table 3, we present the results of our proposed deep learning models for both Word2Vec and FastText based word representations, differing in the type of datatsets being used to produce the word embeddings. Overall accuracies of all models are greater when embeddings trained on Hinglish plus English data, rather than just Hinglish data are used. One possible reason for this observation can be the additional coverage of semantics and coorelations between the word vectors of English data, which can be used for code mixed Hinglish data, thus providing additional knowledge and serving as prior information for Hinglish embeddings data. The process works analogous to a knowledge transfer step in which embeddings for English data are used as prior knowledge for embeddings of Hinglish data. Moreover, Word2Vec embeddings produce better results than FastText embeddings, for all the models. One major reason for this observation is the presence of code mixed data which does not allow character n-grams to be the primary criteria for classification, in the case of FastText embeddings, since the character n-grams belong to the constructs of two different languages. Due to the same reason, context based word vectors i.e. the Word2Vec representations perform better than the character n-grams representations in case of FastText embeddings.

The lack of clean data and linguistic complexities associated with code-mixed data are the major challenges related to the task of sarcasm detection in Hindi-English code mixed data. To allow the model to accommodate the noise in textual data, spelling errors, multiple contexts, and stemming words, even larger data is required along with cautiously labelled classes.

\section{Conclusion}

Social media, in recent years, has become a medium widely used by people for expression of thoughts and opinions, further leading to the realisation of tasks like emotion analysis and opinion mining. Sarcastic content in these texts make it even more challenging to figure out the overall sentiment of the text, thus needing proper processing and analysis.

In this paper, we presented a class-balanced Hindi-English code mixed dataset for the problem of sarcasm detection, by scraping relevant tweets from twitter. We compared two representations, FastText and Word2Vec, both based on different word representation learning mechanisms and trained on custom scraped data from scratch. We created two versions of embeddings, one trained with purely Hinglish data, the other with a mixture of Hinglish and English data, and compared the performance in each case. We analyzed the performance of different deep learning models, which take as input the generated word embeddings, to solve the problem of sarcasm detection. As future work, we plan to compare the vectors aligned with multilingual word embeddings after generation using MUSE with FastText pre aligned word embeddings. We can also explore BERT embeddings and evaluate their performance on the same task.

\bibliographystyle{acl_natbib}
\bibliography{emnlp2020}

\end{document}